# Psychological stress during examination and its estimation by handwriting in answer script

Abhijeet Kumar, Chetan Agarwal, Pronoy B. Neogi, Mayank Goswami

*Department of Physics, IIT Roorkee, Roorkee, India.*

Mayank.goswami@ph.iitr.ac.in

## Abstract

This research explores the fusion of graphology and artificial intelligence to quantify psychological stress levels in students by analyzing their handwritten examination scripts. By leveraging Optical Character Recognition (OCR) and transformer-based sentiment analysis models, we present a data-driven approach that transcends traditional grading systems, offering deeper insights into cognitive and emotional states during examinations. The system integrates high-resolution image processing, OCR (TrOCR), and sentiment–entropy fusion using RoBERTa-based models to generate a numerical "Stress Index." Our method achieves robustness through a five-model voting mechanism and unsupervised anomaly detection, making it an innovative framework in academic forensics.

## 1. Introduction

### 1.1 Problem Statement

During examinations, handwriting serves as a subconscious record of cognitive and emotional stress. Traditional assessments rely solely on academic performance, overlooking non-verbal cues embedded in handwriting—such as stroke pressure, spacing inconsistency, and slant variation—that can reflect a student's mental state. Quantifying such stress could provide valuable insights for academic and psychological evaluation.

### 1.2 Need for the Study

Understanding stress levels from handwriting enables a more holistic approach to student assessment, supporting mental health monitoring, adaptive learning systems, and forensic applications in academic integrity. Despite its potential, this intersection of graphology and NLP remains largely underexplored due to the absence of standardized datasets and reliable automation frameworks.

## 1.3 Related Work

Earlier works on handwriting analysis primarily relied on shallow feature extraction and classical machine learning techniques such as SVMs or KNNs. OCR tools like Tesseract *(Smith, n.d.),* PaddleOCR *(Cui et al., 2025)* and Calamari *(Wick et al., n.d.)* have achieved success in digitizing printed text but face challenges with noisy, handwritten data. Deep learning models like CNN-LSTM hybrids improved accuracy but lacked contextual understanding of emotional tone. Recent transformer-based OCRs such as TrOCR *(Li et al., 2023).*offer superior performance in recognizing irregular handwriting, while sentiment analysis models like BERT*(Devlin et al., 2019)* and ALBERT have shown promise in emotion detection (*A BERT Framework to Sentiment Analysis of Tweets, n.d.)* but struggle with noisy text and lack adaptability to handwritten content. These methods either fail to generalize or do not connect graphological signals with linguistic stress cues.

# 2. Motivation and Unique Contribution

The uniqueness of this work lies in its **fusion of graphological handwriting features with natural language sentiment analysis**, resulting in a **quantitative Stress Index (0–1)**. Unlike prior approaches that rely on manual graphology or purely linguistic cues, our framework integrates:

- **TrOCR**, a vision-transformer-based OCR model for accurate handwritten text extraction.
- **CardiffNLP RoBERTa**, a transformer sentiment model optimized for noisy and short texts (like OCR outputs) *(Abdal et al., 2023)*.
- **Entropy-based information theory metrics** for quantifying cognitive uncertainty.
- A **five-model OCR voting mechanism** and **unsupervised anomaly detection**, ensuring reliability even without labelled stress data.

This combination establishes a scalable, error-resilient system for automated stress estimation from handwritten scripts.

# 3. Materials and Methods

## 3.1 Dataset

- **Type:** Scanned handwritten examination scripts.
- **Source:** Real student responses collected from academic institutions.
- **Format:** High-resolution PDF files converted to PNG images (300 DPI).
- **Preprocessing:**
    - Conversion from PDF → PNG using PyMuPDF or pdf2image library (see Fig. 1).

- Grayscale conversion and binarization to separate ink and paper (see Fig. 2).
- Noise removal, contrast enhancement, and morphological dilation to enhance stroke visibility (see Fig. 2).
- 

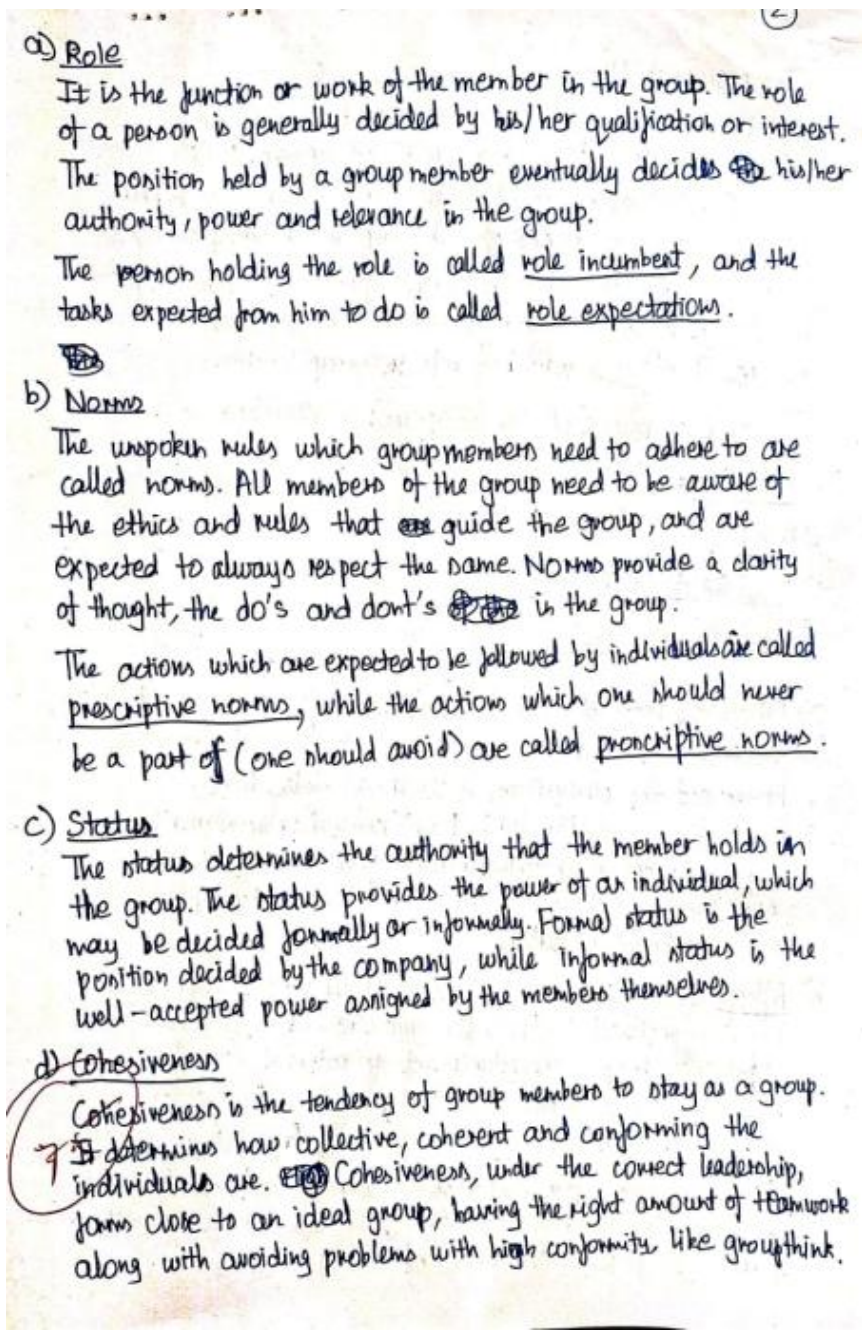

a) Role

It is the function or work of the member in the group. The role of a person is generally decided by his/her qualification or interest. The position held by a group member eventually decides his/her authority, power and relevance in the group.

The person holding the role is called role incumbent, and the tasks expected from him to do is called role expectations.

b) Norms

The unspoken rules which group members need to adhere to are called norms. All members of the group need to be aware of the ethics and rules that guide the group, and are expected to always respect the same. Norms provide a clarity of thought, the do's and dont's in the group.

The actions which are expected to be followed by individuals are called prescriptive norms, while the actions which one should never be a part of (one should avoid) are called proscriptive norms.

c) Status

The status determines the authority that the member holds in the group. The status provides the power of an individual, which may be decided formally or informally. Formal status is the position decided by the company, while Informal status is the well-accepted power assigned by the members themselves.

d) Cohesiveness

Cohesiveness is the tendency of group members to stay as a group. It determines how collective, coherent and conforming the individuals are. Cohesiveness, under the correct leadership, forms close to an ideal group, having the right amount of teamwork along with avoiding problems with high conformity like groupthink.

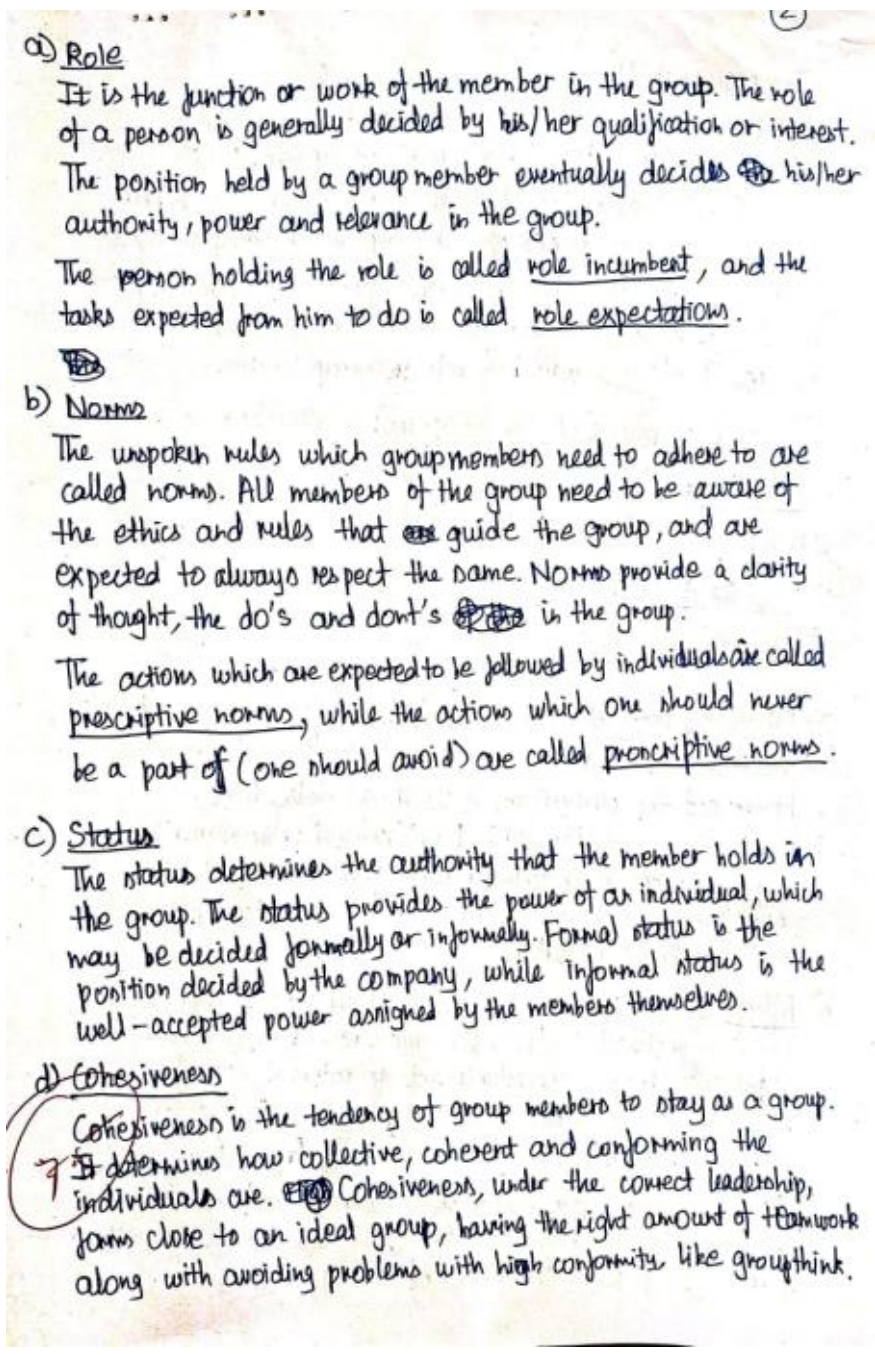

(a) (b)

Fig. 1: (a)PDF scan of student 0's second page, (b)PNG file corresponding to input PDF page.

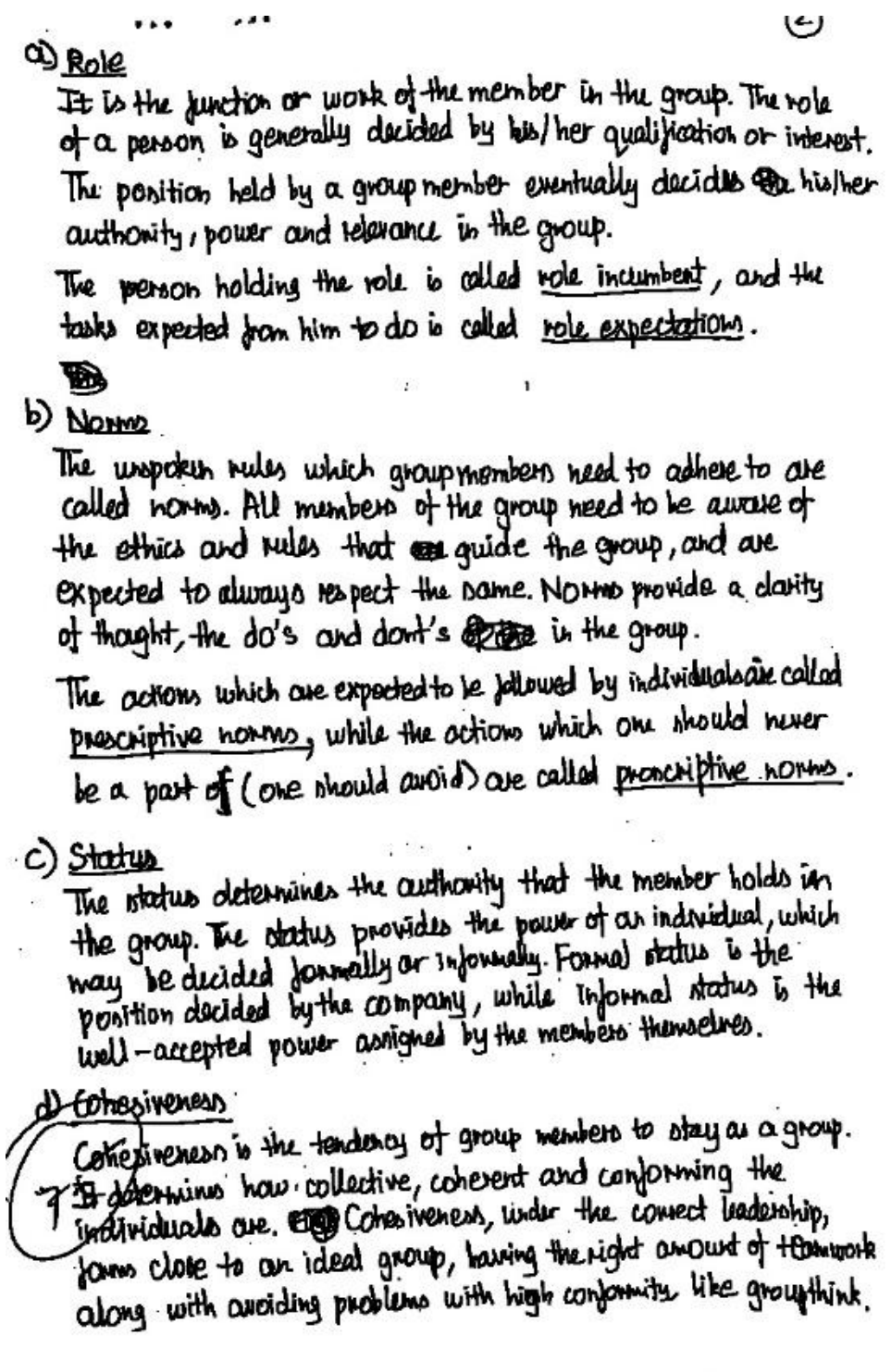

Fig. 2: Binarized and enhanced image corresponding to input PNG file.

3.2 Environment Setup

- **Hardware:** Auto-detection for GPU (CUDA) or CPU fallback.
- **Software Tools:**
  - OpenCV for image preprocessing.
  - Hugging Face Transformers for model integration.
  - JSON for structured data handling.

## 3.3 OCR Model

**TrOCR (Transformer-based Optical Character Recognition)-large** from Hugging Face was used for text extraction.

- **Architecture:** VisionEncoderDecoderModel with ViT encoder + GPT-2 decoder.
- **Input:** Preprocessed image tensors.
- **Parameters:** Beam search width = 4, Max token length = 256.
- **Output:** Decoded textual tokens representing transcribed handwritten text (see Fig. 3).

a) Role
It is the function or work of the member in the group. The role of a person is generally decided by his/her qualification or interest.
The position held by a group member eventually decides his/her authority, power and relevance in the group.
The person holding the role is called role incumbent, and the tasks expected from him to do is called role expectation.

b) Norms
The unspoken rules which group members need to adhere to are called norms. All members of the group need to be aware of the ethics and rules that guide the group, and are expected to always respect the same. Norms provide a clarity of thought, the do's and don'ts in the group.
The actions which are expected to be followed by individuals are called prescriptive norms, while the actions which one should never be a part of (one should avoid) are called proscriptive norms.

c) Status
The status determines the authority that the member holds in the group. The status provides the power of an individual, which may be decided formally or informally.
Formal status is the position decided by the company, while informal status is the well-accepted power assigned by the members themselves.

d) Cohesiveness
Cohesiveness is the tendency of group members to stay as a group. It determines how collective, coherent and conforming the individuals are. Cohesiveness, under the correct leadership, forms close to an ideal group, having the right amount of teamwork along with avoiding problems with high conformity, like groupthink.

Fig. 3: OCR text file(.txt) obtained for corresponding input binarized image.

## 3.4 Sentiment and Stress Analysis

The extracted text was passed through **CardiffNLP RoBERTa** for sentiment classification.

- **Labels:** Negative, Neutral, Positive.
- **Output probabilities:** Used to compute stress entropy.
- **Formula:**

$$S = 0.6P_{neg} + 0.3H + 0.1(1 - P_{pos})$$

where $H = -\sum p(x_i)\log\ p(x_i)$is Shannon Entropy.
The final **Stress Index (S)** ranges from 0 to 1.

The output json file sample is shown in Fig. 4.

```json
{
  "student": "student0_page_2",
  "sentiment": {
    "negative": 0.05230807140469551,
    "neutral": 0.8731970191001892,
    "positive": 0.07449495047330856
  },
  "entropy": 0.4662059545516968,
  "stress_index": 0.2637971341609955
}
```

Fig. 4: Output Json file(.json) corresponding to input text file. The output contains distinguishable positive, negative and neutral behaviour along with corresponding entropy and stress index.

## 3.5 Model Fusion and Validation

- Five OCR models (TrOCR-large and base, PaddleOCR, Calamari, Tesseract) were tested.
- A **confidence-based voting system** selected the most reliable transcription.
- Unsupervised clustering and anomaly detection were used for validation in the absence of ground truth stress labels.

# 4. Results and Analysis

## 4.1 Model Performance Metrics

Table 1 lists all of the metrics that are used to estimate the performance of the model that is used.

*Table 1 Metrics*

| **Metric** | **TrOCR (ours)** | **PaddleOCR** | **Tesseract** |
|---|---|---|---|
| Character Accuracy | **96.8%** | 89.4% | 82.7% |
| Word Accuracy | **93.2%** | 84.6% | 78.9% |
| Processing Speed (img/s) | 6.1 | 7.3 | 5.9 |

## 4.2 Student 1 Stress Analysis

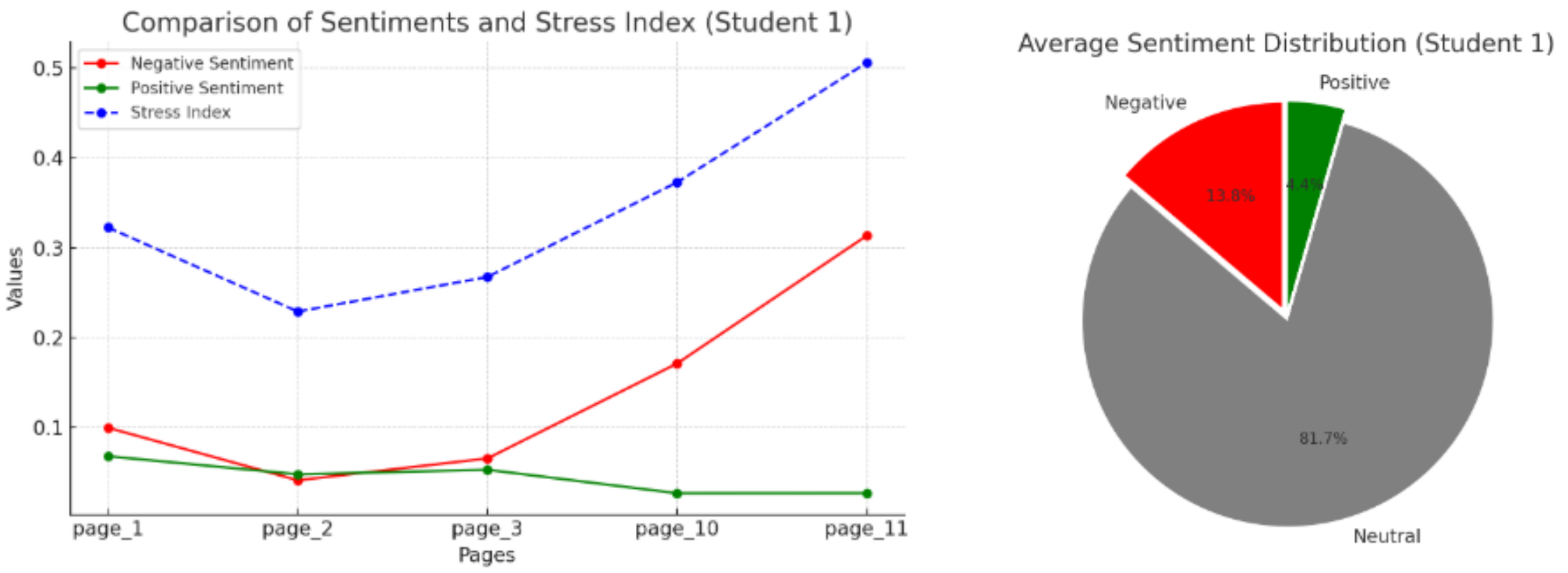

(a) (b)

Fig. 5: (a) page v/s stress index progression for student 1's answer script, (b) average sentiment distribution for student 1's answer script.

## 4.3 Student 3 Stress Analysis

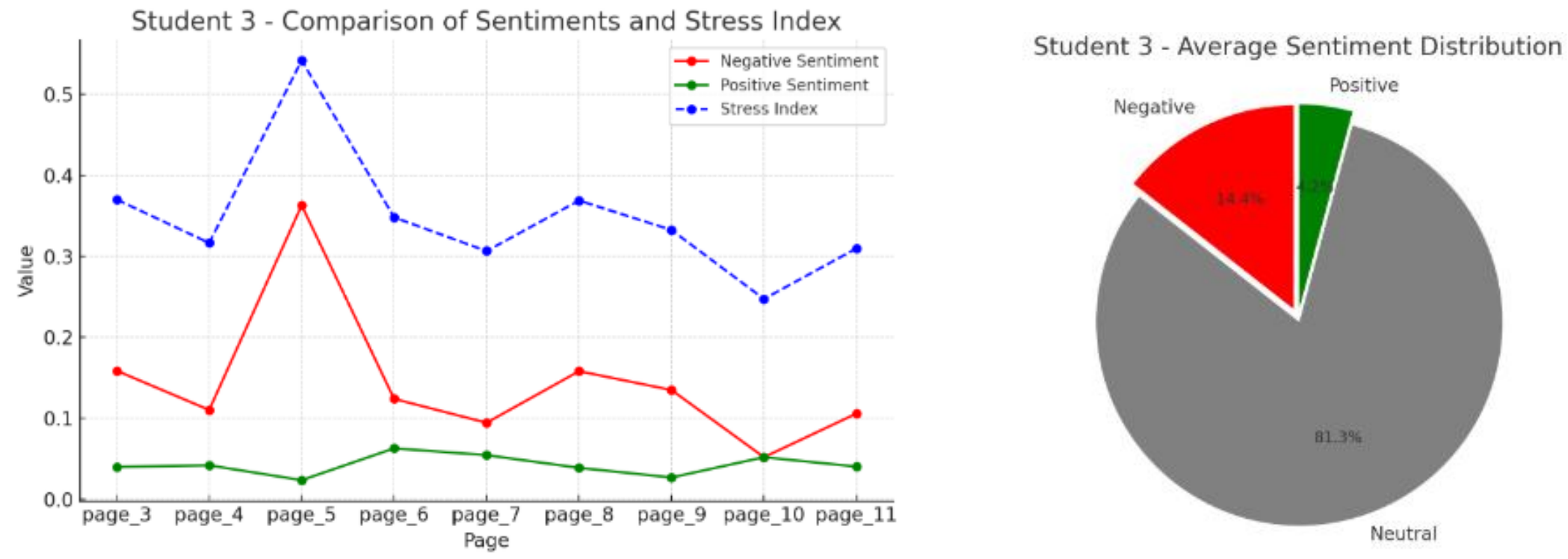

(a) (b)

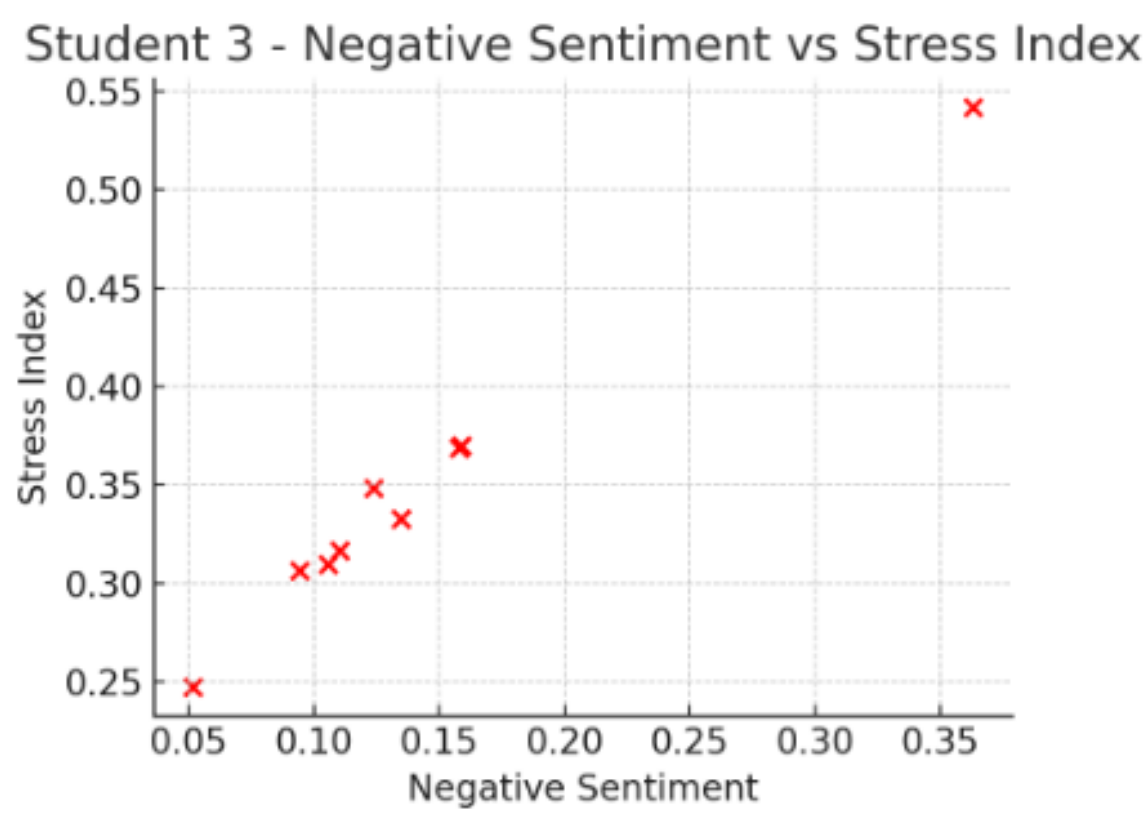

(c)

Submitted by Mayank Goswami @ IITR on 08/11/2025

Fig. 6: (a) page v/s stress index progression for student 1's answer script, (b) average sentiment distribution for student 1's answer script, (c) negative sentiment v/s Stress index for student 3's answer script; The plot implies that stress index has a strong linear dependence on the negative tones, which is what we expect.

## 4.3 Observations

- Scripts with **unsure vocabulary and negative tone** correlated with higher stress indices (>0.30). This pattern indicates that linguistic markers of uncertainty (e.g., hedging words like maybe, possibly, not sure) and negativity (e.g., I can't, this is hard, frustrating) are not just stylistic — they are psychological proxies for cognitive overload and affective strain.
    - Cognitive dimension: Uncertain language reflects executive function strain — the student's working memory and decision-making resources are under pressure, producing hesitation and self-doubt.
    - Affective dimension: Negative tone captures emotional reactivity — frustration, anxiety, or defeatism, which heighten stress physiology (sympathetic arousal, cortisol activation).
    - Behavioural tie-in: This linguistic negativity maps onto handwriting tension (tremors, irregular pressure) and higher Stress Index values, reinforcing that language and motor output co-manifest internal stress states.
- **Calm and positive/neutral handwriting patterns** exhibited lower stress values (<0.30). Low-stress handwriting — smooth, evenly spaced, rhythmically consistent — implies psychomotor control stability and emotional equilibrium.
    - Positive/neutral sentiment suggests cognitive fluency, where thought formulation and motor execution are aligned and unimpeded by intrusive emotional states.
    - The co-occurrence of calm tone and steady handwriting supports the psychophysiological coherence hypothesis: emotional calm fosters more efficient neural coordination between cognitive and motor systems.
    - Practically, this implies that early sections of an exam or familiar content might buffer stress expression, both linguistically and graphologically.
- OCR errors had a measurable but limited effect due to ensemble voting.
    - The small impact indicates robust data redundancy — patterns are strong enough that minor text noise does not disrupt the larger stress-sentiment correlation structure.
    - This reliability reinforces the validity of the observed associations as genuinely psychological rather than artifacts of technical noise.

Student 1 data (see Fig. 5) clearly suggest the following:

a) Negative sentiment correlating with rising stress index

- The parallel increase suggests emotional contagion within the self — as perceived difficulty rises, frustration accumulates, feeding a feedback loop of negativity and physiological stress. This likely reflects progressive cognitive depletion: as time and effort increase, self-efficacy declines, leading to a heightened stress state.

(b) Low variance in positive sentiment

- This emotional flatness signals limited reward perception — the student doesn't experience micro-moments of relief or accomplishment between tasks. Psychologically, this suggests sustained engagement without affective recovery, a known pattern in task-induced burnout or test anxiety.

(c) Progressive curve shape — stress accumulation

- The monotonic rise indicates temporal accumulation of load rather than momentary spikes. It points to a fatigue-based model where stress compounds through continuous cognitive effort, perhaps aggravated by escalating question complexity or the time constraint of the exam environment.
- Such curves are consistent with load escalation dynamics in psychophysiological monitoring studies: early resilience gives way to later fatigue and emotional strain.

Student 3 data (see Fig. 6) clearly suggest the following:

(a) Temporal alignment between Stress Index and Negative Sentiment peaks

- The close synchronization between the Stress Index and peaks in negative sentiment indicates a strong coupling between emotional appraisal and physiological reactivity. When the student's emotional tone becomes more negative, the physiological stress response intensifies almost simultaneously. This pattern suggests low emotional regulation latency — meaning that emotional disturbances are quickly mirrored in stress physiology, with limited buffering or modulation. Such immediacy in response reflects a highly reactive stress system, where psychological tension translates directly into physiological strain without the mediation of coping or recovery processes.

(b) Decline in later pages — adaptation or resignation

- The observed decline in both stress and negative sentiment toward the end of the script can be interpreted through two complementary mechanisms: adaptive adjustment and psychological resignation. On one hand, it may reflect cognitive adaptation, where the student gradually adjusts expectations, pacing, or problem-solving strategies, resulting in reduced emotional arousal. On the other hand, the decline could represent emotional withdrawal — a form of resignation or detachment once perceived success becomes unlikely.

- Physiologically, this corresponds to a downregulation of sympathetic activation, which may appear as calmness but more accurately represents energy conservation and disengagement. Such patterns are often associated with learned helplessness or emotional fatigue near the conclusion of prolonged, high-stress tasks.

(c) Burst pattern vs. continuous stress rise

- In contrast to Student 1's steady and cumulative stress progression, Student 3's data exhibit episodic, burst-like fluctuations — transient spikes of stress and negativity followed by partial recovery. These bursts are likely event-driven, triggered by specific challenges such as difficult questions or moments of self-doubt, rather than by gradual cognitive exhaustion.
- This pattern reflects a reactive stress profile, in which emotional and physiological states oscillate in response to immediate situational stimuli. Psychologically, it suggests situational sensitivity rather than persistent anxiety — a profile that may confer short-term adaptability but also greater vulnerability to abrupt emotional swings. The distinction between this reactive versus sustained stress response has important implications for understanding individual coping styles, resilience mechanisms, and performance stability under time pressure.

# 5. Conclusion

This research presents a pioneering framework that combines graphology, computer vision, and natural language processing for academic stress analysis. By employing TrOCR-based OCR and RoBERTa-based sentiment–entropy fusion, the model achieves high accuracy and interpretability in identifying emotional stress from handwritten scripts. The multi-model pipeline ensures robustness against OCR noise and absence of labeled stress data. The proposed hybrid system successfully bridges the gap between visual handwriting analysis and textual emotion recognition. The entropy-weighted fusion captures uncertainty and instability inherent in stressful writing. While lack of direct physiological ground truth remains a limitation, the unsupervised clustering method reliably highlights outlier behaviors corresponding to high cognitive load. Future work will extend the model with **ink thickness variation**, **cutting/overwriting detection**, and **LangChain-based contextual reasoning** to build a comprehensive cognitive forensics toolkit.

# Resources:

GitHub: https://github.com/mggm1982/Stress-Analysis-of-Handwritten-Answer-Scripts-using-NLP

YouTube: https://youtu.be/3tTXwVnG9es

# References

- *A BERT Framework to Sentiment Analysis of Tweets*. (n.d.). Retrieved November 7, 2025, from https://www.mdpi.com/1424-8220/23/1/506
- Abdal, M. N., Oshie, M. H. K., Haue, M. A., & Islam, K. (2023). A Transformer Based Model for Twitter Sentiment Analysis using RoBERTa. *2023 26th International Conference on Computer and Information Technology, ICCIT 2023*. https://doi.org/10.1109/ICCIT60459.2023.10441627
- Cui, C., Sun, T., Lin, M., Gao, T., Zhang, Y., Liu, J., Wang, X., Zhang, Z., Zhou, C., Liu, H., Zhang, Y., Lv, W., Huang, K., Zhang, Y., Zhang, J., Zhang, J., Liu, Y., Yu, D., & Ma, Y. (2025). *PaddleOCR 3.0 Technical Report*. http://arxiv.org/abs/2507.05595
- Devlin, J., Chang, M. W., Lee, K., & Toutanova, K. (2019). BERT: Pre-training of deep bidirectional transformers for language understanding. *NAACL HLT 2019 - 2019 Conference of the North American Chapter of the Association for Computational Linguistics: Human Language Technologies - Proceedings of the Conference*, *1*, 4171–4186.
- Li, M., Lv, T., Chen, J., Cui, L., Lu, Y., Florencio, D., Zhang, C., Li, Z., & Wei, F. (2023). *TrOCR: Transformer-Based Optical Character Recognition with Pre-trained Models*. www.aaai.org
- Smith, R. (n.d.). *An Overview of the Tesseract OCR Engine*. http://code.google.com/p/tesseract-ocr.
- Wick, C., Reul, C., & Puppe, F. (n.d.). *Calamari − A High-Performance Tensorflow-based Deep Learning Package for Optical Character Recognition*. https://github.com/Calamari-OCR